\documentclass[journal]{IEEEtran}

\usepackage[vlined, ruled, linesnumbered, commentsnumbered]{algorithm2e}
\usepackage{amsmath}
\usepackage{cite}
\usepackage{color}
\definecolor{chred}{rgb}{0.8,0,0}
\definecolor{chgray}{rgb}{0.5,0.5,0.5}
\usepackage{graphicx}
\usepackage{caption}
\usepackage{dblfloatfix}
\usepackage{subfigure}
\usepackage{eqlist}
\usepackage{txfonts}
\usepackage{url}
\usepackage{footmisc}
\usepackage{booktabs}

\DeclareMathOperator*{\argmax}{arg\, max}
\usepackage{multirow}
\begin{document}
\title{Assembly Sequence Planning for Motion Planning}

\author{Weiwei~Wan,~\IEEEmembership{Member,~IEEE,}
        Kensuke~Harada,~\IEEEmembership{Member,~IEEE,}
        Kazuyuki~Nagata,~\IEEEmembership{Member,~IEEE}
\thanks{Weiwei Wan, Kensuke Harada, and Kazuyuki Nagata are with National
Institute of Advanced Industrial Science and Technology (AIST), Japan. Kensuke
Harada is also affiliated with Osaka University, Japan.
{\tt\small wan-weiwei@aist.go.jp}}}

\markboth{Journal of \LaTeX\ Class Files,~Vol.~x, No.~x, xxxx~2016}%
{Shell \MakeLowercase{\textit{et al.}}: Bare Demo of IEEEtran.cls for IEEE Journals}

\maketitle

\begin{abstract}
This paper develops a planner to find an optimal assembly sequence to
assemble several objects. The input to the planner is the mesh models of the
objects, the relative poses between the objects in the assembly, and the final
pose of the assembly. The output is an optimal assembly sequence,
namely (1) in which order should one assemble the objects, (2) from which
directions should the objects be dropped, and (3) candidate grasps of
each object. The proposed planner finds the optimal
solution by automatically permuting, evaluating, and searching the possible assembly sequences
considering stability, graspability, and assemblability qualities.
It is expected to guide robots to do assembly using
translational motion. The output provides initial and goal
configurations to motion planning algorithms. It is ready to be used by robots
and is demonstrated using several simulations and real-world executions.
\end{abstract}


\IEEEpeerreviewmaketitle

\section{Introduction}

\IEEEPARstart{A}{ssembly} planning implies a wide range of concepts. It includes
task and symbolic planning in the high level, motion planning in the middle level, and
force and torque control in the low level. In this paper, we focus on the
high-level assembly sequence planning problem and develop a planner that could
automatically find an optimal assembly sequence which is ready to be
used by robots for motion planning. The input to the planner
includes

\begin{itemize}
  \itemsep0em
  \item Mesh model of a robotic hand.
  \item Mesh models of objects.
  \item Relative poses between objects in the assembly.
  \item Goal pose of the assembly.
\end{itemize}

\noindent The output includes 

\begin{itemize}
  \itemsep0em
  \item Assembly order: Which objects to assemble first.
  \item Assembly direction: How to drop or insert objects.
  \item Accessible grasps: How to grasp objects during assembly.
\end{itemize}

The proposed planner finds the optimal solution by automatically
permuting, evaluating, and searching the possible assembly sequences
considering stability, graspability, and assemblability qualities.
It is expected to guide robots to do assembly using
translational motion. The planner is born for motion planning as the output
provides initial and goal configurations for robots to carry out motion
planning algorithms.

\begin{figure}[!htbp]
	\centering
	\includegraphics[width=.46\textwidth]{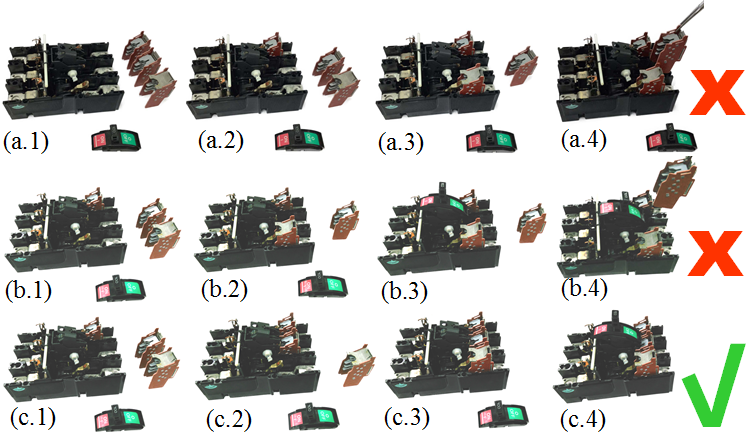}
	\caption{Assembly a switch. (a) and (b) are two bad choices due to collision
	between grippers and surrounding capacitors or collision between the active
	capacitor and the finished part. (c) is an optimal sequence.}
	\label{teaser}
\end{figure}

The study is motivated by an assembly task in real world where the goal is to
assemble a switch shown in Fig\ref{teaser}(c.4). The switch is
composed of five parts shown in Fig.\ref{teaser}(a.1). A robot needs to insert
three capacitors into a base and attach a switch button on top of it. An
optimal sequence to finish this task is, as was described in the last sentence,
to insert the three capacitors first and then attach the switch button on top of
it. Fig.\ref{teaser}(c.1-4) illustrates this optimal solution. In contrast, a
bad assembly sequence is shown in Fig.\ref{teaser}(a.1-4) where after inserting
two surrounding capacitors in (a.2) and (a.3), it is difficult to find an collision-free grasp
to insert the third capacitor into the middle slot (see (a.4)). Another
bad assembly sequence is shown in Fig.\ref{teaser}(b) where two capacitors are
inserted in (b.1) and (b.2), and the switch button is attached in (b.3). It
is impossible to insert the third capacitor in (b.4).
Human beings could easily find inserting the third capacitor in (a.4) and (b.4)
are bad choices since the surrounding capacitors or the switch button would
block the motion. However, it is a non-trivial problem to robots.
Traditionally methods used in robotic assembly are that skilled human
technicians teach robots the assembly orders and directions, which makes robotic
manufacturing less robotic.

The planner challenges the non-trivial problem by performing assembly sequence
planning. It finds the optimal assembly sequence by automatically permuting,
evaluating, and searching the objects. First, it permutes
the objects and lists all assembly orders. Each assembly order includes a
sequence of objects that should be assembled sequentially. Then, for each
assembly order, the planner manipulate the object in the sequence one by one
and evaluate the stability and the graspability of each manipulated object.
Meanwhile, it checks if the manipulated object can be assembled, computes its
optimal assembly directions considering the normals of contact surfaces, and evaluates the assemblability
(tolerance to errors) of the optimal assembly directions.
Using these permuting, evaluating, and searching steps, the approach is able to
find some optimal assembly orders and directions that are (1) stable after
assembling each manipulated object, (2) have lots of accessible grasps and are
flexible to the kinematic constraints of robots, and (3) robust to assembly
errors.

Comparing with contemporary studies, our main contribution is we do everything
automatically in 3D workspace, considering not only stability and
assemblability, but also graspability. The benefit is the planned results
could be seamlessly used by robot motion planners: The
accessible grasps work as initial and goal
configurations for robot end-effectors; The assembly directions work as
motion primitives. The assembly planner is born for motion planning. 
The experimental section of this paper not only presents and analyzes the
results of the assembly planner, but also includes some real-world executions
that use integrated assembly sequence planning and motion planning.



\section{Related Work}%

Early studies in assembly planning are symbolic reasoning systems
\cite{Mello90sim}\cite{Defazio87sim} and use given contact and assembly
constraints to do decision search. For example, Mello, et al. \cite{Mello91sim}
is a representative one which used logical expressions to defined the assemblies
and used a relation model graph to generate assembly sequences. It considered geometric-feasibility,
mechanical-feasibility, and stability during searching.
Sanderson \cite{Sanderson99sim} performed robust symbolic assembly planning
by considering the clearance between contacting objects.
It is the first work which used the keyword ``assemblability''.
Reference \cite{Thomas01sim} is a more recent work  that used symbolic
reasoning. The work is quite practical as it included some real-world
executions. Knepper, et al.
\cite{Knepper13sim} also ran some real-world executions. The study not only
used symbolic planning to find assembly sequences, but also used geometric analysis
to infer how to attach pins to holes. In another paper \cite{Knepper14sim},
Knepper further formulated assembly sequence planning as a task allocation
problem and proposed a method to plan assembly sequences that could be done in parallel.

Comparing with symbolic assembly planning, geometric reasoning systems
generate assembly sequences by automatically discovering contact and
assembly constraints.
The earliest geometric reasoning-based assembly planning systems we could find
is \cite{Woo91sim}. The work used geometric constraints to build a disassembly
tree and employed the tree to find an assembly sequence for a restricted class of
problems. Another representative study is reference \cite{Wilson94sim} which
used the geometric constraints to build a Non-Directional Blocking Graph (NDBG) and
employed the graph to reason about assembly sequences. Romeny, et al.
\cite{Romney95sim} extended the NDBG to 3D assembly by using a sphere of
directions of motion. One region on the sphere was associated with one DBG and
the sphere and the DBGs together added up to the NDBG. Assembly sequences were
planned by analyzing the sphere and its associated DBGs. Ikeuchi, et al.
\cite{Ikeuchi94sim} used constraint Gaussian spheres to represent the
contact constraints between mesh models, and planned assembly sequences by
considering the constraint spheres at each contact. Thomas, et al.
\cite{Thomas03sim} represented the assembled objects using stereographical
projections and figured out a different way to denote separability. The
stereographical projection essentially shares the same idea with constraint
spheres except that it could handle complex polyhedrons quickly. A good summary
of the studies that plan assembly sequences using geometric reasoning before
2013 could be found in \cite{Jimenez13sim}. 

The early assembly planning systems only considered a single constraint model
(mostly geometric constraints).
More recent work considers a mixed model of constraints.
For example, 
Agrawala, et al. \cite{Agrawala03sim} used NDBG to generate assembly sequence
and used visibility constraints to find an view-friendly assembly sequence which
could be drawn on a paper document as assembly instructions. The constraint
model is a mixture of geometric constraints and visibility constraints.
Ostrovsky-Berman, et al.
\cite{OB06sim} discussed the tolerance of different contact types and used them
to optimize assembly sequence planning. The tolerance is similar to the concept
of ``assemblability'' in \cite{Sanderson99sim}, and the constraint is a mixture
of geometric constraints and uncertainty constraints. Schwarzer, et al.
\cite{Schwarzer10sim} additionally considered $m$-handed assembly which allowed
$m$ objects to be disassembled simultaneously.
Wei \cite{Wei12sim} applied the automatic assembly sequence generation to ship
building by considering the sizes, positions, and materials of the objects.
Dobashi, et al. \cite{Dobashi14sim} additionally considered the collision-free
grasps between manipulated objects and the assembled objects during assembling,
although the assembly sequence is pre-defined manually considering these
constraints.
McEvoy, et al. \cite{McEvoy14sim} considered both stability and geometric
constraints in planning the assembly sequences of truss structures. A real-world
execution is included in their paper. Dogar, et al. \cite{Dogar15sim} used several mobile robots
to assemble a chair. The constraints between robots, between robot
grippers and the assembled objects, and between manipulated object and assembled
objects, are considered during the assembly. The paper also includes a
real-world execution. Ghandi, et al. \cite{Ghandi15sim} made a good summary of the
studies dealing various constraints before 2015.

In this paper, we perform assembly sequence planning considering
statics constraints of the assembled part (stability),
quasistatic constraints between grippers and manipulated
objects and geometric constraints between grippers and surrounding objects
(graspability), and geometric constraints between the manipulated object and the
finished part (assemblability).
We evaluate the quality of stability, grasplability, and assemblability to
find an optimal sequence for translational assembly. We assume the
objects are 3D polyhedron, the assembly motion are translational, and a single gripper is
used at one time. Comparing with previous work, we not only present planners
which consider a mixed model of the constraints that directly relate to robot
motion planning, but also demonstrate the pragmatic flavor of our system using
the real-world executions of several exemplary tasks.

\section{Overview of the Approach}

\begin{figure}[!htbp]
	\centering
	\includegraphics[width=.46\textwidth]{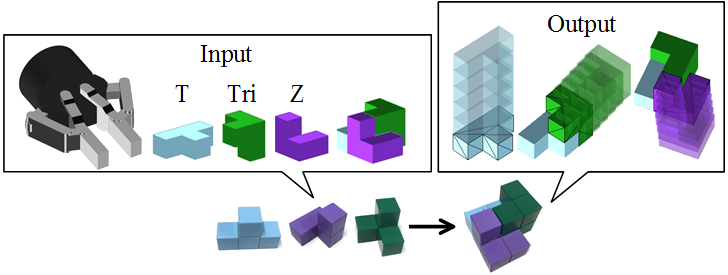}
	\caption{The input and output of the planner.}
	\label{inout}
\end{figure}

We present the algorithmic part of this paper using soma cube as an example to
promote clarity. Soma cube is a solid dissection puzzle invented in 1933. Three
blocks, an z-shape block, a t-shape block, and a tri-shape block of the soma
cube puzzle are used to assemble a given structure (see Fig.\ref{inout}). The
input to the planner is the mesh models of the three objects, their relative
poses, and the final pose of the assembly. The output is the assembly order, the
assembly directions, and the candidate grasps of each object.
The input and output are shown in the frameboxes of Fig.\ref{inout} (The
candidate grasps are not shown in the figure).
  
\begin{figure}[!htbp]
	\centering
	\includegraphics[width=.48\textwidth]{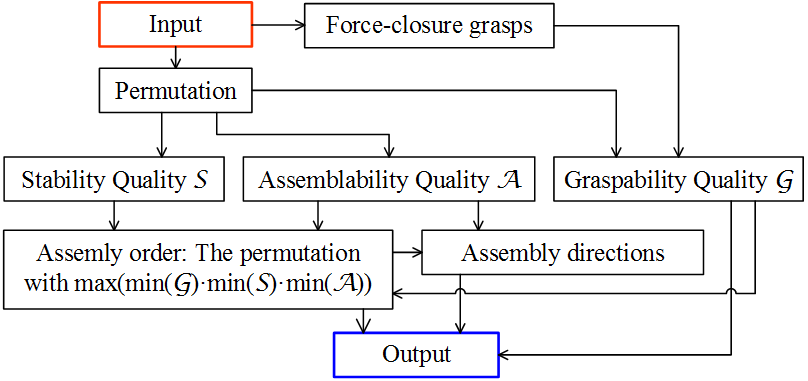}
	\caption{The algorithmic flow of the proposed approach.}
	\label{algflow}
\end{figure}

The algorithmic flow of the proposed approach is shown in Fig.\ref{algflow}. The
input includes: (1) The mesh model of the robotic hand; (2) The mesh models of
the objects; (3) The relative poses between the objects in the assembled structure;
(4) The final pose of the assembly. First, using the number of objects, the
planner computes all possible permutations in the "permutation" toolbox. For
each permutation, the approach evaluates its stability, graspability, and
assemblability qualities. The stability quality is evaluated by computing the
relationship between $\mathbf{p}_{com}$ (center of mass) and the
boundary of supporting area. It is denoted by $\mathcal{S}$ in the figure.
The graspaibility quality is evaluated by computing the number
force-closure and collision-free grasps. Using the mesh model of the robotic
hand and the mesh models of the objects, the planner computes the possible hand
configurations to grasp the object in the ``force-closure grasps'' box without
considering collisions with other objects. The ``graspability quality'' box
removes the force-closure grasps that collide with the finished part and counts
the number of remaining grasps as the quality of graspability. Graspability is
denoted by $\mathcal{G}$ in Fig.\ref{algflow}. The assemblability quality is
evaluated using the normals of the contact faces between the manipulated object
and the finished part. The process is done in the ``Assemblability Quality''
box. If the current permutation is assemblable, the planner
uses the direction that has largest clearance from all contact normal as the
assembly direction, and sets its quality $\mathcal{A}$ considering the size of the clearance.
After evaluating the qualities, the approach compares the $\mathcal{G}$,
$\mathcal{S}$, and $\mathcal{A}$ of each permutation, selects the permutation that has
$\mathtt{max}(\mathtt{min}(\mathcal{G})\cdot\mathtt{min}(\mathcal{S})\cdot
\mathtt{min}(\mathcal{A}))$ as the optimal assembly order, and selects its
correspondent assembly directions as the optimal assembly direction. The details
of this expression will be explained in next section.

\section{Implementation Details}


\subsection{Permutation}

Given the total number of the objects to be assembled, permutation permutes
object IDs and lists all the possible assembly orders without considering any
constraints. $n$ objects lead to $P_n^n=n!$ permuted orders. There are three
cubes in the example shown in Fig.\ref{inout} and therefore the number of
permuted orders is $3!=6$. Consider three objects with ID ``Z'', ``T'', ``Tri''
(Fig.\ref{inout}), the output of the permutation is $\textnormal{Tri} \leftarrow
\textnormal{Z} \leftarrow \textnormal{T}$, $\textnormal{Tri} \leftarrow \textnormal{T} \leftarrow
\textnormal{Z}$, $\textnormal{Z} \leftarrow \textnormal{Tri} \leftarrow
\textnormal{T}$, $\textnormal{Z} \leftarrow \textnormal{T} \leftarrow
\textnormal{Tri}$, $\textnormal{T} \leftarrow \textnormal{Z} \leftarrow
\textnormal{Tri}$, $\textnormal{T} \leftarrow \textnormal{Tri}
\leftarrow \textnormal{Z}$, and each element, for example $\textnormal{T} \leftarrow
\textnormal{Tri} \leftarrow \textnormal{Z}$, indicates a potential assembly
order which first assembles object $\textnormal{Tri}$ to $\textnormal{T}$, and
then assembles object $\textnormal{Z}$ to the complex of $\textnormal{T}$ and
$\textnormal{Tri}$.
For a subsequence $\textnormal{T} \leftarrow \textnormal{Tri}$, $\textnormal{T}$
is called the base object, $\textnormal{Tri}$ called the manipulated object.
When assembling $\textnormal{Z}$ to the complex of $\textnormal{T}$ and
$\textnormal{Tri}$, $\textnormal{Z}$ is the manipulated object.
$(\textnormal{T},\textnormal{Tri})$ is the base.

The reason permutation is used instead of AND/OR graph is we not only consider
assembling two objects using ``AND'', but also consider the order of the
assembly, namely which one is the base and which one is the manipulated object.
(Note: Some potential assembly orders may be infeasible. They will be removed
progressively when computing the qualities.)

\subsection{Stability}

\begin{figure}[!htbp]
	\centering
	\includegraphics[width=.48\textwidth]{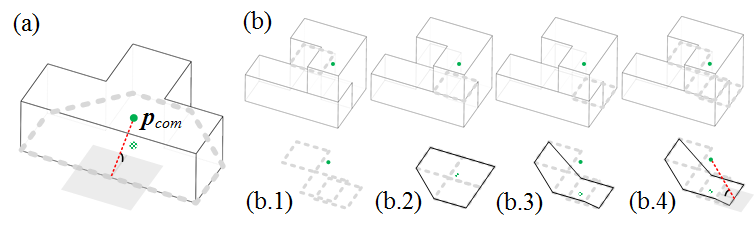}
	\caption{Computing the stability of the manipulated object by measuring the
	angle between $\protect\overrightarrow{\mathbf{p}_{b}\mathbf{p}_{com}}$ and a
	horizontal plane.
	(a) shows the first step. The red dash line shows the vector
	$\protect\overrightarrow{\mathbf{p}_{b}\mathbf{p}_{com}}$. The grey square
	indicates a horizontal plane. (b) shows the second step where the supporting boundary is 3D.}
	\label{stability}
\end{figure}

Stability is evaluated sequentially for each object in each
potential order. The first step is check if the assembled part is
stable after assembling the manipulated object following a given potential
order. The stability qualities of unstable objects will be set to 0. Deciding
whether an object is stable can be performed by projecting its center of mass
$\mathbf{p}_{com}$ and supporting area to a horizontal plane, and checking if
the projected $\mathbf{p}_{com}$ is inside the convex hull of the projected
area. The object is not stable the projected $\mathbf{p}_{com}$ is outside the
hull. The green shadow point and the dash boundaries in Fig.\ref{stability}(a)
and (b.3) are the projections of $\mathbf{p}_{com}$ and the supporting area. The
two manipulated objects are both stable.

If the manipulated object is stable, the next step is to evaluate
its stability quality. This is implemented by finding the nearest point
$\mathbf{p}_{b}$ on the convex boundary of its supporting area (which might be
from both the base and the environment) to the manipulated object's
$\mathbf{p}_{com}$ and compute the angle between the vector
$\overrightarrow{\mathbf{p}_{b}\mathbf{p}_{com}}$ and the horizontal plane. A
smaller angle avoids large disturbance torques, indicating higher
stability.

Take $\textnormal{T}\leftarrow\textnormal{Tri}\leftarrow\textnormal{Z}$, for example. The first
step of stability evaluation is to compute the stability of T. Since
the final configuration of the assembly is a given parameter, the
poses of $\textnormal{T}$, $\textnormal{Tri}$, and $\textnormal{Z}$ are pre-known. The first step therefore equals to evaluating the stability of
T at a given pose. Fig.\ref{stability}(a) shows this simple case
and the angle that indicates the quality.
The second step is to evaluate the stability of object Tri. Computing the angle
between $\overrightarrow{\mathbf{p}_{b}\mathbf{p}_{com}}$ and the horizontal
plane becomes complicated since the supporting area
could be from both the table surface and the surface of T. We solve
this problem by using the 3D boundary shown in
Fig.\ref{stability}(b) and (b.1, b.3, b.4).

The third step evaluates the stability of object Z. Like the second object, the
third object could also be supported by both the table surface and the surfaces
of the finished part. We use the same technique as Tri to compute its stability
quality.

The result of the stability evaluation is a triple $\textit{\textbf{s}}$=($s_1$,
$s_2$, $s_3$) where each element indicates the stability quality of each object when doing assembly
following the potential order. The stability qualities of all permuted orders
form a column of triples named $\mathcal{S}$ where

\begin{equation}
\mathcal{S}=\begin{pmatrix}
\textit{\textbf{s}}_1\\
\textit{\textbf{s}}_2\\
\ldots\\
\textit{\textbf{s}}_6\\
\end{pmatrix}=\begin{pmatrix}
s_{11} & s_{12} & s_{13}\\
s_{21} & s_{22} & s_{23}\\
\ldots\\
s_{61} & s_{62} & s_{63}\\
\end{pmatrix}
\end{equation}

\subsection{Graspability}

\begin{figure}[!htbp]
	\centering
	\includegraphics[width=.48\textwidth]{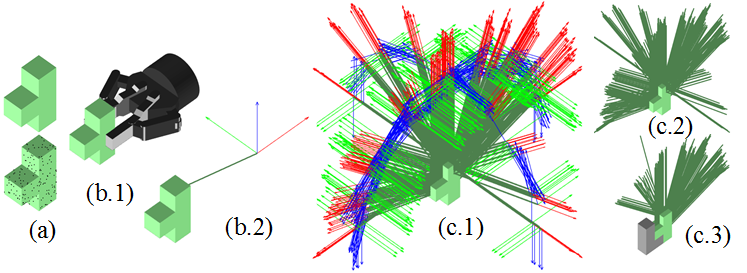}
	\caption{Computing the graspability of the manipulated object by counting the
	number of accessible grasps. (a) Sample the surface of the object model. (b.1)
	For each pair of the sample, find force-closure grasps. (b.2) One grasp is
	represented by a segment plus a coordinate attached to its end. (c.1) All
	grasps after rotating and collision checking with the object itself and the
	table surface. (c.2) Simplified representation (coordinates are not shown). If
	this object is the first object, the number of segments in (c.2) will be its
	graspability. If the object is not the first one ((c.3)), remove the grasps
	that collide with the finished part (grey object in (c.3)) and count the remaining
	grasps as the graspability.}
	\label{graspability}
\end{figure}

For a potential order computed by (1), its graspability is computed by
sequentially counting the force-closure and collision-free grasps of each
object. The process is the same as a precedent work where we compute the
force-closure grasps of all candidate objects (Section 4 of
\cite{Wan2016arsim}). During assembly planning, we
remove the collided grasps from the precomputed force-closure set and count the
number of remaining grasps (known as accessible grasps) as the graspability.
An example is shown in Fig.\ref{graspability}.
For the first object in the potential order, we only check the collision between the
hand and the table surface. For the second and third objects, we check both the
collision between the hand and the table, and the collision between the hand and
the finished part. The output of graspability evaluation for a potential order
is a triple $\textit{\textbf{g}}$=($g_1$, $g_2$, $g_3$) where each element
indicates the graspability quality of each object when doing assembly following the order. The graspability
qualities of all permuted orders form a column of triples named $\mathcal{G}$ where

\begin{equation}
\mathcal{G}=\begin{pmatrix}
\textit{\textbf{g}}_1\\
\textit{\textbf{g}}_2\\
\ldots\\
\textit{\textbf{g}}_6\\
\end{pmatrix}=\begin{pmatrix}
g_{11} & g_{12} & g_{13}\\
g_{21} & g_{22} & g_{23}\\
\ldots\\
g_{61} & g_{62} & g_{63}\\
\end{pmatrix}
\end{equation}

\subsection{Assembly directions and assemblability}

The assembly directions and assemblability of a potential assembly order are
computed and evaluated using the normals of the contact faces
between the newly added object and the finished part. In theory, we
are using the constraint spheres shown in the second row of
Fig.\ref{assembledirections}.
In implementation, we compute the convex hull of the contact normals and perform
piece-wise analysis considering the types of the convex hull and the position
of the origin with respect to the hull (third row of
Fig.\ref{assembledirections}). For each object, we compute
its optimal assembly direction $\textbf{\textit{n}}_o$ and set its
assemblability quality to a certain value considering the clearance of
$\textbf{\textit{n}}_o$ from surrounding constraints (purple arrows and
numerical values in the third row of
Fig.\ref{assembledirections}).
The following cases are considered in the implementation:

\begin{figure*}[!htbp]
	\centering
	\includegraphics[width=.98\textwidth]{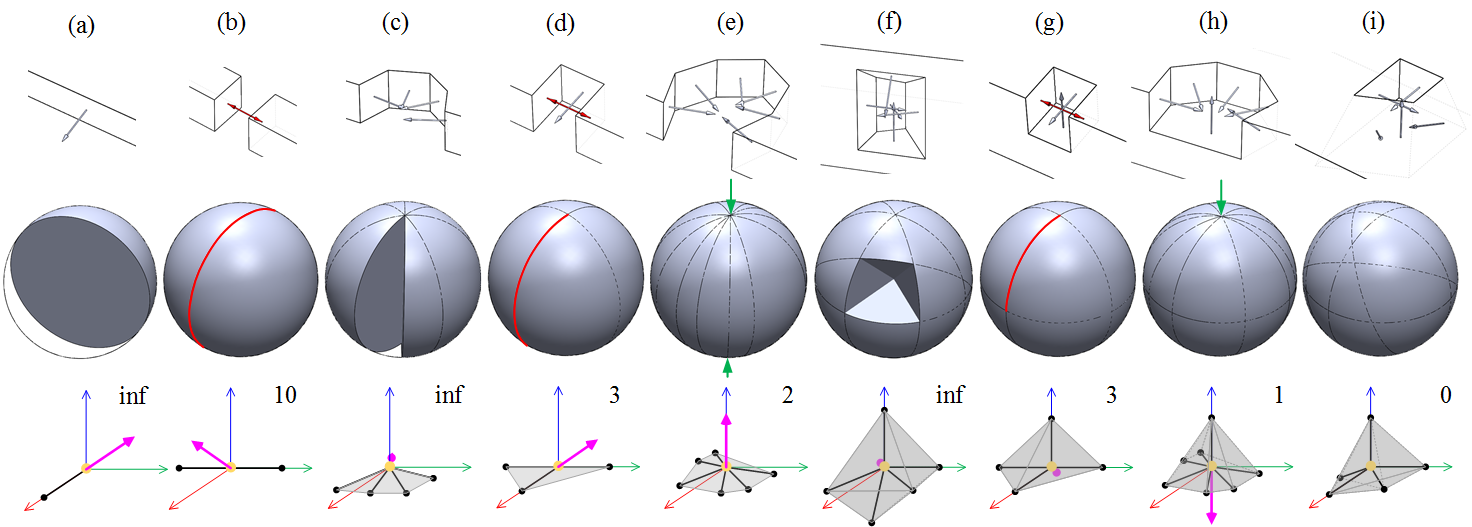}
	\caption{Computing and evaluating the assembly directions and assemblability
	using piece-wise analysis considering the types of the convex hull and the position
	of the origin. The first row shows the different types of contacts and the
	contact normals. The second row show the correspondent constraint spheres. The
	third row are the convex hulls in the space of contact normals. The purple
	arrows in the third row show the chosen optimal assembly directions. The
	numbers are their assemblability qualities.}
	\label{assembledirections}
\end{figure*}

\subsubsection{Fig.\ref{assembledirections}(a)}
The convex hull of the contact normal is a single vector.
This is the simplest case. There could be one or more contact faces but the
contact normals are the same. The first row of Fig.\ref{assembledirections}(a)
shows an example with only one contact face. The supplementary cone of the contact normal is a
hemisphere shown in the second row of (a). The manipulated object can
approach the base from the directions that are not blocked by the hemisphere.
In the space of contact normals, the convex hull is a vector (third row).

For this case, we choose the contact normal as the optimal assembly direction
$\textbf{\textit{n}}_o$ and assign an infinite assemblability value to it
(infinity indicates high assemblability). The purple arrow and the
numerical value in the third row of Fig.\ref{assembledirections}(a) show the
chosen $\textbf{\textit{n}}_o$ and its assemblability quality. The optimal
direction is essentially the normal of the blocked hemisphere.
It has the largest clearance from being blocked.

\subsubsection{Fig.\ref{assembledirections}(b)}
The convex hull of the contact normal is a line passing the origin.
There could be two or more contact faces but the contact normals are along two
opposite directions. The first row of
Fig.\ref{assembledirections}(b) is an example of this case. The supplementary
cone are composed of two opposite hemispheres. The manipulated object can
approach the base from any direction on the plane squeezed by the two
hemispheres (marked in red in Fig.\ref{assembledirections}(b)).
Suppose the first contact normal is $\textit{\textbf{n}}_1$,
we randomly choose one direction from
$\textbf{\textit{n}}_x$, $\textbf{\textit{n}}_x\cdot\textbf{\textit{n}}_1=0$ as
$\textbf{\textit{n}}_o$ and set its assemblability value to 10 (a
relatively large value). The purple arrow and the
number in the third row of Fig.\ref{assembledirections}(b) show the
chosen $\textbf{\textit{n}}_o$ and its assemblability quality.

\subsubsection{Fig.\ref{assembledirections}(c)}
The convex hull of the contact normal is a 2D polygon and the origin is on one
vertex of the polygon. There are at least two contact faces and all the contact
normals are on the same plane. The first row of Fig.\ref{assembledirections}(c)
shows an example.
The supplementary cone is the union of several hemispheres. The remaining part of
the constraint sphere is a spherical wedge. The manipulated object can
approach the base from any direction inside the wedge.
Suppose the normals are
$\textit{\textbf{n}}_1$, $\textit{\textbf{n}}_2$, \ldots,
$\textit{\textbf{n}}_n$, we use the normalized value of
$\textit{\textbf{n}}_1$+$\textit{\textbf{n}}_2$+\ldots$\textit{\textbf{n}}_n$
as $\textbf{\textit{n}}_o$ and assign an infinite assemblability value to it
(like Fig.\ref{assembledirections}(a), the assembly is quite flexible). The
chosen $\textbf{\textit{n}}_o$ and its assemblability quality are illustrated by
a purple arrow and a number in the third row of
Fig.\ref{assembledirections}(c).

\subsubsection{Fig.\ref{assembledirections}(d)}
The convex hull of the contact normal is a 2D polygon and the origin is on one
edge of the polygon. There is at least one pair faces whose normals are
opposite and the normals of all faces are on the same plane. If there is more
than one pair of opposite faces, their normals must be parallel with the first
pair. The first row of Fig.\ref{assembledirections}(d) shows an example with one
pair of opposite faces. The supplementary cone is the union of two opposite
hemispheres and some cross hemispheres. The remaining part of the constraint
sphere is a half plane (marked in red in Fig.\ref{assembledirections}(d)).
The manipulated object can approach the base from any direction on the half
plane.
Suppose $(\textit{\textbf{n}}_j$, $(\textit{\textbf{n}}_k)$ is one of
the opposite pair, we use the normalized value of
$\hat{\sum{\textit{\textbf{n}}_i}}\cdot(\textbf{1}-\textit{\textbf{n}}_j)$,
where
$\textit{\textbf{n}}_i\cdot\textit{\textbf{n}}_j\neq\pm1$, as $\textbf{\textit{n}}_o$.
Here, $\hat{\sum{\textit{\textbf{n}}_i}}\cdot(\textbf{1}-\textit{\textbf{n}}_j)$
indicates the projection of $\hat{\sum{\textit{\textbf{n}}_i}}$ on the plane
perpendicular to $\textit{\textbf{n}}_j$. The assemblability value is set to 3
(a medium value). The purple arrow and the
number in the third row of Fig.\ref{assembledirections}(d) show the
chosen $\textbf{\textit{n}}_o$ and its assemblability quality.

\subsubsection{Fig.\ref{assembledirections}(e)}
The convex hull of the contact normal is a 2D polygon and the origin is inside
the polygon. There are at least three contact faces and all the contact normals
are on the same plane. It is different from Fig.\ref{assembledirections}(c) in
that the contact faces nearly form a circle and block lateral insertion. The
first row of Fig.\ref{assembledirections}(e) is an example.
The supplementary cone of the contact normals covers the whole sphere except the
directions along the polar lines (marked using green arrows in
Fig.\ref{assembledirections}(e)).
The manipulated object can approach the base along the two polar directions.
Suppose there are two contact normals $\textit{\textbf{n}}_j$ and
$(\textit{\textbf{n}}_k$ where
$\textit{\textbf{n}}_j\cdot(\textit{\textbf{n}}_k\neq1$, we
choose $\textit{\textbf{n}}_j\times\textit{\textbf{n}}_k$ as
$\textbf{\textit{n}}_o$ and set its value to 2 (a relatively small value). The
purple arrow and the number in the third row of
Fig.\ref{assembledirections}(e) show the chosen $\textbf{\textit{n}}_o$ and its
assemblability quality.

\subsubsection{Fig.\ref{assembledirections}(f)}
The convex hull of the contact normal is a 3D polyhedron and the origin is on
one vertex of the polyhedron. There are at least three contact faces whose
normals are not in the same plane. The first row of
Fig.\ref{assembledirections}(f) is an example. The supplementary cone of the
contact normals cross each other like the middle figure of
Fig.\ref{assembledirections}(f). The remaining part of the supplementary cone is
a sphere sector and the manipulated object can approach the base from any
direction inside the sector. Suppose the normals are
$\textit{\textbf{n}}_1$, $\textit{\textbf{n}}_2$, \ldots,
$\textit{\textbf{n}}_n$, we use the normalized value of
$\textit{\textbf{n}}_1$+$\textit{\textbf{n}}_2$+\ldots$\textit{\textbf{n}}_n$
as $\textbf{\textit{n}}_o$ and assign an infinite assemblability value to
it (like Fig.\ref{assembledirections}(a) and (c), the assembly is quite
flexible).The chosen $\textbf{\textit{n}}_o$ and its assemblability quality are
illustrated by a purple arrow and a number in the third row of
Fig.\ref{assembledirections}(f).

\subsubsection{Fig.\ref{assembledirections}(g)}
The convex hull of the contact normal is a 3D polyhedron and the origin is on
one edge of the polyhedron. There is at least one pair faces whose normals are
opposite and at least two other contact normals that are neither parallel with
the opposite normals nor opposite. If there is more
than one pair of opposite faces, their normals must be parallel with the first
pair. The first row of
Fig.\ref{assembledirections}(e) is an example. Like
Fig.\ref{assembledirections}(d), the supplementary cone is the union of two
opposite hemispheres plus some cross hemispheres. The remaining part of the constraint
sphere is a sector (marked in red in Fig.\ref{assembledirections}(e)). The
manipulated object can approach the base from any direction inside the sector. 
Like Fig.\ref{assembledirections}(d),
suppose $(\textit{\textbf{n}}_j$, $(\textit{\textbf{n}}_k)$ is one of
the opposite pair, we use the normalized value of
$\hat{\sum{\textit{\textbf{n}}_i}}\cdot(\textbf{1}-\textit{\textbf{n}}_j)$,
where $\textit{\textbf{n}}_i\cdot\textit{\textbf{n}}_j\neq\pm1$, as
$\textbf{\textit{n}}_o$. The assemblability value is set to 3 (a medium value).
The purple arrows and numbers in the third row of
Fig.\ref{assembledirections}(g) shows the chosen $\textbf{\textit{n}}_o$ and its assemblability quality.

\subsubsection{Fig.\ref{assembledirections}(h)}
The convex hull of the contact normal is a 3D polyhedron and the origin is on
one face of the polyhedron. This is an extension of the case in
ig.\ref{assembledirections}(e). There are at least three contact faces and all
the contact normals are on the same plane. Meanwhile, there is at least one
extra normal that crosses the plane. The first row of
Fig.\ref{assembledirections}(h) is an example. The manipulated object can only
approach the base along one polar direction
(marked using a green arrow in Fig.\ref{assembledirections}(h)).
The face where the origin locates has two normals, and we use the normal that
points outside the polyhedron as $\textbf{\textit{n}}_o$. The assemblability value is
set to 1 (a small value). The purple arrows and numbers in the third row of
Fig.\ref{assembledirections}(g) shows the chosen $\textbf{\textit{n}}_o$ and its assemblability quality.

\subsubsection{Fig.\ref{assembledirections}(i)}
The convex hull of the contact normal is a 3D polyhedron and the origin is
inside the polyhedron. In this case, the supplementary cone of the contact
normals cover the whole sphere and there is no way to perform the assembly. The
assemblability value is set to 0.

\subsubsection{Collisions along $\textbf{\textit{n}}_o$}
The results computed in cases \textit{1)}-\textit{9)} are not final. See
Fig.\ref{collisionsonno}(a) for example. The goal is to assemble
the green block into the hole. Following Fig.\ref{assembledirections}(h),
$\textbf{\textit{n}}_o$ will be the purple arrow shown in
Fig.\ref{collisionsonno}(a). However, this direction is infeasible since the
green block will collide with a handle of the base as it moves along
$\textbf{\textit{n}}_o$. A feasible solution is shown in
Fig.\ref{collisionsonno}(b) and Fig.\ref{collisionsonno}(c) which requires
inserting the block into the handle and assembling the block
from a nearer spot. Planning the motion in
Fig.\ref{collisionsonno}(b)$\rightarrow$(c) is not an assembly planning problem.
It is a motion planning problem which is still unsolved (see ``narrow
passages'' in \cite{Kavraki05sim}).

The assembly planner proposed in this paper is designed for motion planning and
would like to avoid leaving this difficulty to motion planning algorithms. 
We therefore define a ``starting offset'' where a robot will start
assembling the manipulated object along $\textbf{\textit{n}}_o$. We compute a
swept volume of the manipulated object translating from the ``starting
offset'' to its goal pose along $\textbf{\textit{n}}_o$, and check if there is
collision between the swept volume and the finished part. Collision indicates
assembling the manipulated object along $\textbf{\textit{n}}_o$ is in feasible
and we reset its assemblability quality to 0.
The resetted assemblability qualities, instead of the the original ones from
\textit{1)}-\textit{9)}, will be used as the final results.

\begin{figure}[!htbp]
	\centering
	\includegraphics[width=.48\textwidth]{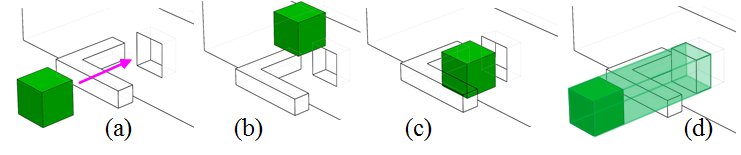}
	\caption{Checking the collisions along the assembly direction. (a)
	The manipulated object (green) will collide with the base. (b, c) It is a
	``narrow passage'' problem in motion planning. (d) Swept volume along the
	assembly direction is used to detect collisions. Collided assembly directions
	will removed to avoid leaving the ``narrow passage'' problem to motion
	planning.}
	\label{collisionsonno}
\end{figure}

The output of assemblability evaluation is a triple in the form of
(($\textbf{\textit{n}}_{o1}$, $a_1$), ($\textbf{\textit{n}}_{o2}$, $a_2$),
($\textbf{\textit{n}}_{o3}$, $a_3$)) where $\textbf{\textit{n}}_{o1}$,
$\textbf{\textit{n}}_{o2}$, and $\textbf{\textit{n}}_{o3}$ indicate the
optimal assembly directions of each object when doing assembly following the
potential order, $a_1$, $a_2$, and $a_3$ are the assemblability qualities of the
assembly directions. The assemblability
qualities of all permuted orders form a column of triples named $\mathcal{A}$.
The assemblability directions forms a accompanying column of triples named
$\mathcal{A}'$.

\begin{equation}
\mathcal{A}=\begin{pmatrix}
\textit{\textbf{a}}_1\\
\textit{\textbf{a}}_2\\
\ldots\\
\textit{\textbf{a}}_6\\
\end{pmatrix}=\begin{pmatrix}
a_{11} & a_{12} & a_{13}\\
a_{21} & a_{22} & a_{23}\\
\ldots\\
a_{61} & a_{62} & a_{63}\\
\end{pmatrix},~
\mathcal{A}'=\begin{pmatrix}
\textbf{\textit{n}}_{1o1} & \textbf{\textit{n}}_{1o2} &
\textbf{\textit{n}}_{1o3}\\
\textbf{\textit{n}}_{2o1} & \textbf{\textit{n}}_{2o2} &
\textbf{\textit{n}}_{2o3}\\
\ldots\\
\textbf{\textit{n}}_{6o1} & \textbf{\textit{n}}_{6o2} &
\textbf{\textit{n}}_{6o3}\\
\end{pmatrix}
\end{equation}

\subsection{Wrap up}
The planner finds the optimal assembly order and
assembly directions by evaluating a mixed model of stability, graspability, and
assemblability qualities. Each quality is expressed as a tuple. The planner
picks the smallest value in each quality and uses the multiplication of the
three smallest values as the overall quality of a potential order.
Mathematically, it is expressed as
$\mathtt{min}(\textit{\textbf{s}})\cdot\mathtt{min}(\textit{\textbf{g}})\cdot
\mathtt{min}(\textit{\textbf{a}})$. The potential order that has largest overall
quality, namely
$\mathtt{max}(\mathtt{min}(\mathcal{G})\cdot\mathtt{min}(\mathcal{S})\cdot
\mathtt{min}(\mathcal{A}))$ is chosen as the optimal order. Here,
$\mathtt{min}()$ computes the smallest value of each row. $\cdot$ indicates
element-wise multiplication. 

\begin{equation}
optimal\_orderid=\argmax_{rowid}
\begin{pmatrix}
\mathtt{min}(\textit{\textbf{s}}_1)\cdot\mathtt{min}(\textit{\textbf{g}}_1)\cdot\mathtt{min}(\textit{\textbf{a}}_1)\\
\mathtt{min}(\textit{\textbf{s}}_2)\cdot\mathtt{min}(\textit{\textbf{g}}_2)\cdot\mathtt{min}(\textit{\textbf{a}}_2)\\
\ldots\\
\mathtt{min}(\textit{\textbf{s}}_6)\cdot\mathtt{min}(\textit{\textbf{g}}_6)\cdot\mathtt{min}(\textit{\textbf{a}}_6)\\
\end{pmatrix}
\end{equation}

The assembly directions that correspond to this
order is selected as the optimal assembly direction:

\begin{equation}
optimal\_directions=\mathcal{A}'(optimal\_orderid, :)
\end{equation}

\subsection{Time Efficiency}
Exploded combinatorics is a big problem during the computation.
In the worst case, the time cost is $O(n!)$ and is NP hard.
However, the worst case seldom appears in the real world thanks to constraints
from the environment (e.g., table surface) and the finished part. To take
advantage of these constraints, we remove the potential orders that are not
stable, have no accessible grasps, and have 0 assemblability qualities
progressively along with the computation of the three qualities. The pseudo code
is shown in Alg.\ref{timeefficiency}. It uses a vector $\mathbf{m}$ to record
the unstable, inaccessible, and inassemblable orders, and avoids recomputing them in new
loops. The inline function $\mathtt{update}()$ updates $\mathbf{m}$ using the
infeasible sub-sequences.

\begin{algorithm}[!htbp]
  \SetKwData{true}{true}
  \SetKwData{false}{false}
  \SetKwFunction{nrows}{nrows}
  \SetKwFunction{ncols}{ncols}
  \SetKwFunction{zeros}{zeros}
  \SetKwFunction{stability}{stability}
  \SetKwFunction{graspability}{graspability}
  \SetKwFunction{assemblability}{assemblability}
  \SetKwFunction{update}{update}
  \SetKwFunction{updateprog}{inline}{}{}
  \DontPrintSemicolon
  \KwData{The permutation $\mathcal{P}$}
  \KwResult{The stability, graspability, and assemblability
  qualities $\mathcal{S}$, $\mathcal{G}$, and $\mathcal{A}$ and $\mathcal{A}'$}
  \Begin {
  	\textit{/*m is the vector to record the infeasible orders*/}\\
  	$n_r$,~$n_c$$\leftarrow$\nrows{$\mathcal{P}$},~\ncols{$\mathcal{P}$}\\
  	$\mathcal{S}$,~$\mathcal{G}$,~$\mathcal{A}$$\leftarrow$\zeros{$n_r$,
  	$n_c$},~$\mathbf{m}$$\leftarrow$[\true]*$n_r$\\
  	\For{$i\in\{1,2,\ldots,n_r\}$}{
  		\If{$\mathbf{m}(i)$}{
	  		\For{$j\in\{1,2,\ldots,n_c\}$}{
	  			$\mathcal{S}(i,j)$$\leftarrow$\stability{$\mathcal{P}(i,j)$}\\
	  			\eIf{$\mathcal{S}(i,j)$==$0$} {
	  				\update{$\mathbf{m}$, $\mathcal{P}(i,1\ldots j)$}, \textbf{break}\\
	  			}
				{
	  				$\mathcal{G}(i,j)$$\leftarrow$\graspability{$\mathcal{P}(i,j)$}\\
		  			\eIf{$\mathcal{G}(i,j)$==$0$} {
	  					\update{$\mathbf{m}$, $\mathcal{P}(i,1\ldots j)$}, \textbf{break}\\
		  			}
		  			{
		  				$\mathcal{A}(i,j)$,~$\mathcal{A}'(i,j)$$\leftarrow$\\~~~~~~~~\assemblability{$\mathcal{P}(i,j)$}\\
			  			\If{$\mathcal{A}(i,j)$==$0$} {
	  						\update{$\mathbf{m}$, $\mathcal{P}(i,1\ldots j)$}\\\textbf{break}\\
			  			}
		  			}
				}
	  		}
  		}
  	}
    \Return{$\mathcal{S}$, $\mathcal{G}$, $\mathcal{A}$}\\
  	\textit{/*Definition of the inline
  	function update()*/}\\
    \updateprog{\update{$\mathbf{m}$, $\mathcal{P}(i,1\ldots j)$}}\\{
    					\For{$k\in\{1,2,\ldots,n_r\}$} {
		  					\If{$\mathcal{P}(k,1\ldots j)$==$\mathcal{P}(i,1\ldots j)$}{
		  						$\textbf{m}(k)$=\false\\
		  					}
		  				}
	}
  }
  \caption{Efficiently evaluating the qualities by progressively removing the
  infeasible assembly orders}
  \label{timeefficiency}
\end{algorithm}

\section{Experiments and Analysis}

\subsection{Soma cubes}

\subsubsection{The problem shown in Fig.\ref{inout}}
\begin{figure}[!htbp]
	\centering
	\includegraphics[width=.48\textwidth]{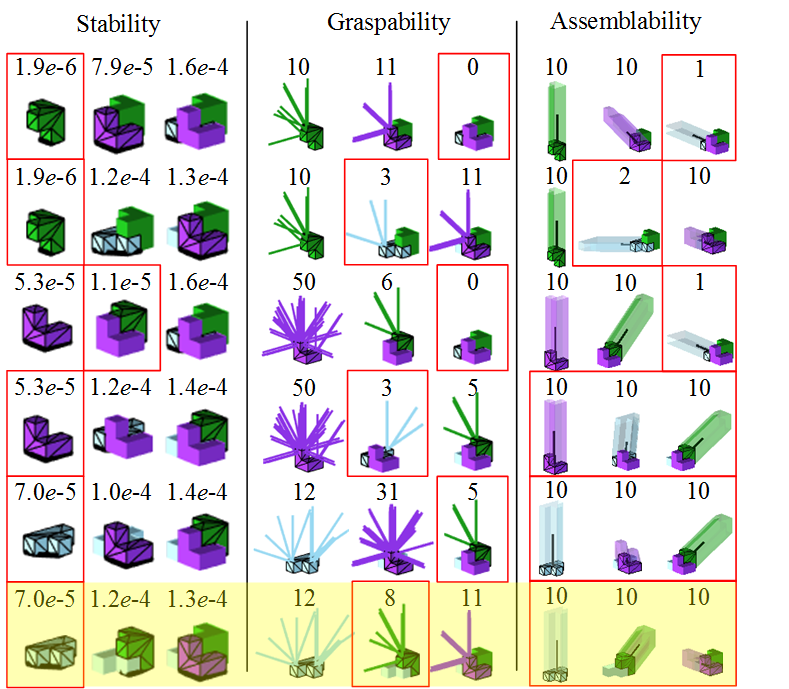}
	\caption{The evaluated stability, graspability, and
	assemblability qualities for all the six potential orders of the exemplary
	problem shown in Fig.\ref{inout}. Each row of a quality column is one potential
	order. The manipulated objects of a potential order are the ones with black
	edges (e.g. the first row of the stability column is Tri$\leftarrow$Z$\leftarrow$T).
	The smallest qualities of each order is marked with red boxes.
	If the objects have the same quality, the whole order is marked.}
	\label{expsoma3}
\end{figure}

Fig.\ref{expsoma3} shows the evaluated stability, graspability, and
assemblability qualities for all the six potential orders of the exemplary
problem shown in Fig.\ref{inout}. Each row is one potential order. The columns
separated by the vertical lines correspond to the three qualities respectively:
The left column is $\mathcal{S}$; The middle column is $\mathcal{G}$; The right column is
$\mathcal{A}$. The optimal order is T$\leftarrow$Tri$\leftarrow$Z and is marked
using yellow shadow. The optimal assembly directions are shown in black segments
in the right column. The accessible grasps are shown using colored segments in
the middle column. The results indicate that the optimal assembling sequence
to assembly the three blocks is as the output in Fig.\ref{inout}.

\begin{figure}[!htbp]
	\centering
	\includegraphics[width=.47\textwidth]{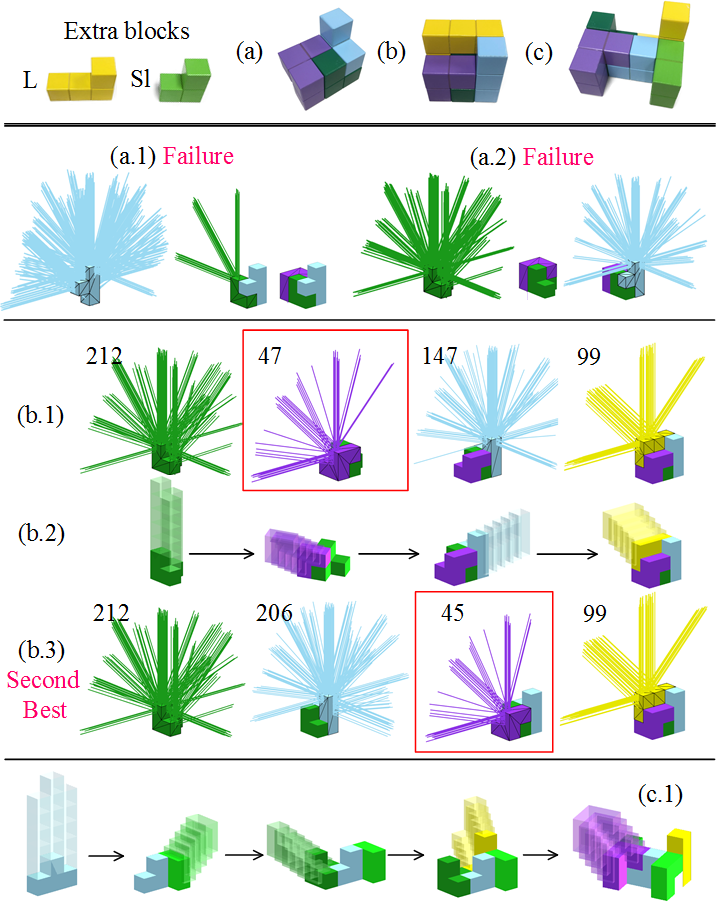}
	\caption{Some other results: (a, a.1, a.2) Three-block assembly with no
	solution; (b, b.1, b.2, b.3) Four-block assembly; (c, c.1, c.2, c.3)
	Five-block assembly.}
	\label{collisionsonno}
\end{figure}

\begin{figure*}[!htbp]
	\centering
	\includegraphics[width=.96\textwidth]{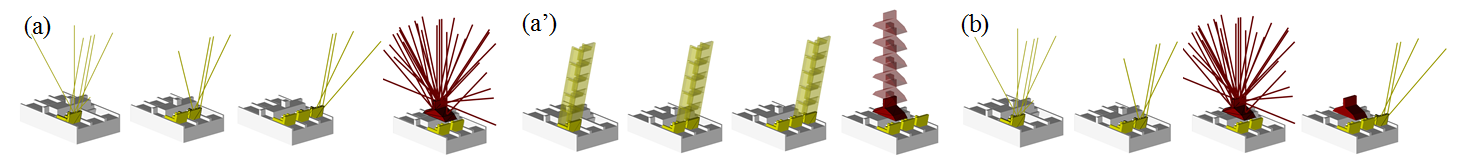}
	\caption{(a, a') The optimal sequence to assemble the switch in
	Fig.\ref{teaser}. (b) Another sequence with the same qualities.}
	\label{expswitch}
\end{figure*}

\subsubsection{Other results using three, four, fifth blocks}
Fig.\ref{collisionsonno} shows some other examples of assembling soma cubes.
Fig.\ref{collisionsonno}(a) uses the same T, Tri, and Z objects, but a different
assembly. It is impossible to assemble them using the hand in
Fig.\ref{collisionsonno}(b.1): Assembling Z before Tri will be unstable;
Assembling Z after Tri leads to zero accessible grasps (the third object in Fig.\ref{collisionsonno}(a.1) and the second step in
Fig.\ref{collisionsonno}(a.2)).
Fig.\ref{collisionsonno}(b) uses an extra L block to assemble a four-block
assembly.
Fig.\ref{collisionsonno}(b.1) and (b.2) is the accessible grasps and optimal
assembly directions of the optimal assembly order. Since we are computing the minimum
$\mathcal{S}$, $\mathcal{G}$, and $\mathcal{A}$ of each assembly order,
the order in Fig.\ref{collisionsonno}(c), which has the same $\mathcal{S}$ and
$\mathcal{A}$ but different $\mathcal{G}$ as the order in
Fig.\ref{collisionsonno}(b.1), will be not be the top choice. The minimum $\mathcal{G}$ of the order in
Fig.\ref{collisionsonno}(c) appears at Z (the third object, the quality is 45).
Although this order has a larger $\mathcal{G}$ at T (the second object, the quality is 206),
it is not as robust. It is more likely to fail since the worst quality is
worse. Fig.\ref{collisionsonno}(c) uses extra L and Sl blocks to
assemble a five-block assembly. An optimal sequence is shown in
Fig.\ref{collisionsonno}(c.1). This optimal sequence is not single. There are
eight other different choices that have the same $\mathtt{max}(\mathtt{min}(\mathcal{G})\cdot\mathtt{min}(\mathcal{S})\cdot
\mathtt{min}(\mathcal{A}))$.

\subsection{Switch}
Fig.\ref{expswitch} shows the planned assembly sequence for the switch shown in
Fig.\ref{teaser}. Although looks complicated, finding the assembly sequence of
the switch is much simplifier than the soma blocks. All values in $\mathcal{A}$
equal 1. Fig.\ref{expswitch}(a) and (a') are the accessible grasps and optimal
assembly directions of the optimal assembly order. Note that there
is another choice that has the same
$\mathtt{max}(\mathtt{min}(\mathcal{G})\cdot\mathtt{min}(\mathcal{S})\cdot
\mathtt{min}(\mathcal{A}))$, which is shown in Fig.\ref{expswitch}(b).

\subsection{Real-world execution}

The planned assembly sequences are sent to robots for motion planning and
execution. Readers are encouraged to refer to \cite{2016arXiv160803140W} for the
details. Results of the real-world execution are in a video attachment.
Some snapshots are as follows.

\begin{figure}[!htbp]
	\centering
	\includegraphics[width=.48\textwidth]{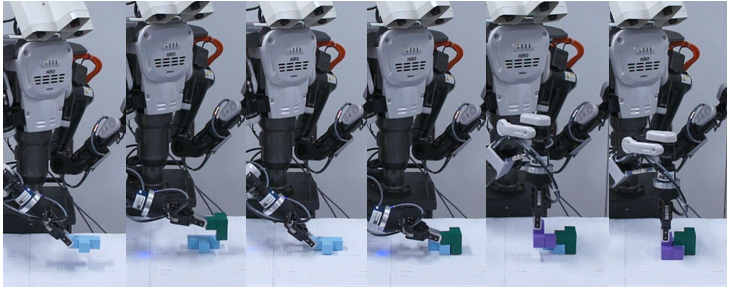}
	\caption{Real-world execution (video attachment).}
	\label{execution}
\end{figure}

\section{Conclusions and Future Work}
An assembly planner is presented in this paper to plan an optimal sequence for
translational assembly. The planner is demonstrated using both soma cube
structures consisting of three, four, and five blocks, and an industry switch.
The planned sequences are used by real robots to do integrated assembly sequence
planning and motion planning, which demonstrates that the assembly sequence
planner could be seamlessly used by robot motion planners.

\section*{Acknowledgment}
The paper is based on results obtained from a project commissioned by the
New Energy and Industrial Technology Development Organization (NEDO).

\ifCLASSOPTIONcaptionsoff
  \newpage
\fi



\bibliographystyle{IEEEtran}
\bibliography{reference}

\begin{thebibliography}{10}
\providecommand{\url}[1]{#1}
\csname url@samestyle\endcsname
\providecommand{\newblock}{\relax}
\providecommand{\bibinfo}[2]{#2}
\providecommand{\BIBentrySTDinterwordspacing}{\spaceskip=0pt\relax}
\providecommand{\BIBentryALTinterwordstretchfactor}{4}
\providecommand{\BIBentryALTinterwordspacing}{\spaceskip=\fontdimen2\font plus
\BIBentryALTinterwordstretchfactor\fontdimen3\font minus
  \fontdimen4\font\relax}
\providecommand{\BIBforeignlanguage}[2]{{%
\expandafter\ifx\csname l@#1\endcsname\relax
\typeout{** WARNING: IEEEtran.bst: No hyphenation pattern has been}%
\typeout{** loaded for the language `#1'. Using the pattern for}%
\typeout{** the default language instead.}%
\else
\language=\csname l@#1\endcsname
\fi
#2}}
\providecommand{\BIBdecl}{\relax}
\BIBdecl

\bibitem{Mello90sim}
L.~Mello, ``{AND/OR Graph Representation of Assembly Plans},'' \emph{IEEE
  Trans. Robot. Autom.}, 1990.

\bibitem{Defazio87sim}
T.~DeFazio \emph{et~al.}, ``{Simplified Generation of All Mechanical Assembly
  Seuences},'' \emph{IEEE J. Robot. Autom.}, 1987.

\bibitem{Mello91sim}
L.~Mello \emph{et~al.}, ``{A Correct and Complete Algorithm for the Generation
  of Mechanical Assembly Sequences},'' \emph{IEEE Trans. Robot. Autom.}, 1991.

\bibitem{Sanderson99sim}
A.~Sanderson, ``{Assemblability Based on Maximum Likelihood Configuration of
  Tolerances},'' \emph{IEEE Trans. Robot. Autom.}, 1999.

\bibitem{Thomas01sim}
U.~Thomas \emph{et~al.}, ``{A System for Automatic Planning, Evaluation and
  Execution of Assembly Sequences for Industrial Robots},'' in \emph{Proc.
  IROS}, 2001.

\bibitem{Knepper13sim}
R.~Knepper \emph{et~al.}, ``{IkeaBot: An Autonomous Multi-Robot Coordinated
  Furniture Assembly System},'' in \emph{Proc. ICRA}, 2013.

\bibitem{Knepper14sim}
------, ``{Distributed Assembly with AND/OR Graphs},'' in \emph{WS: AI
  Robotics, IROS}, 2014.

\bibitem{Woo91sim}
T.~Woo \emph{et~al.}, ``{Automatic Disassembly and Total Ordering in Three
  Dimensions},'' \emph{Trans. ASME}, 1991.

\bibitem{Wilson94sim}
R.~Wilson \emph{et~al.}, ``{Geometric Reasoning About Mechanical Assembly},''
  \emph{Artif. Intell.}, 1994.

\bibitem{Romney95sim}
B.~Romney \emph{et~al.}, ``{An Efficient System for Geometric Assembly Sequence
  Generation and Evaluation},'' in \emph{Proc. ICEC}, 1995.

\bibitem{Ikeuchi94sim}
K.~Ikeuchi \emph{et~al.}, ``{Toward an Assembly Plan from Observation Part I:
  Task Recognition With Polyhedral Objects},'' \emph{IEEE Trans. Robot.
  Autom.}, 1994.

\bibitem{Thomas03sim}
U.~Thomas \emph{et~al.}, ``{Efficient Assembly Sequence Planning Using
  Stereographical Projections of C-Space Obstacles},'' in \emph{Proc. ISATP},
  2003.

\bibitem{Jimenez13sim}
P.~Jimenez, ``{Survey on Assembly Sequencing: A Combinatorial and Geometrical
  Perspective},'' \emph{J. Intell. Manuf.}, 2013.

\bibitem{Agrawala03sim}
M.~Agrawala \emph{et~al.}, ``{Design Effective Step-by-step Assembly
  Instructions},'' in \emph{Proc. SIGGRAPH}, 2003.

\bibitem{OB06sim}
Y.~Ostrovsky-Berman \emph{et~al.}, ``{Relative Position Computation for
  Assembly Planning With Planar Toleranced Parts},'' \emph{Int. J. Robot.
  Res.}, 2006.

\bibitem{Schwarzer10sim}
F.~Schwarzer and othters, ``{Efficient Linear Unboundedness Testing: Algorithm
  and Applications to Translational Assembly Planning},'' \emph{Int. J. Robot.
  Res.}, 2010.

\bibitem{Wei12sim}
Y.~Wei, ``{Automatic Generation of Assembly Sequence for the Planning of
  Outfitting Process in Shipbuilding},'' Ph.D. dissertation, TU Deft, 2012.

\bibitem{Dobashi14sim}
H.~Dobashi \emph{et~al.}, ``{Robust Grasping Strategy for Assemblying Parts in
  Various Shapes},'' \emph{Adv. Robotics}, 2014.

\bibitem{McEvoy14sim}
M.~McEvoy \emph{et~al.}, ``{Assembly Path Planning for Stable Robotic
  Construction},'' in \emph{Proc. TePRA}, 2014.

\bibitem{Dogar15sim}
M.~Dogar \emph{et~al.}, ``{Multi-Robot Grasp Planning for Sequential Assembly
  Operations},'' in \emph{Proc. ICRA}, 2015.

\bibitem{Ghandi15sim}
S.~Ghandi \emph{et~al.}, ``{Review and Taxonomies of Assembly and Disassembly
  Path Planning Problems and Approaches},'' \emph{Comput. Aided Des.}, 2015.

\bibitem{Wan2016arsim}
W.~Wan \emph{et~al.}, ``{Achieving High Success Rate in Dual-arm Handover Using
  Large Number of Candidate Grasps},'' \emph{Adv. Robotics}, 2016.

\bibitem{Kavraki05sim}
H.~Choset \emph{et~al.}, \emph{{Principles of Robot Motion: Theory, Algorithms,
  and Implementations}}.\hskip 1em plus 0.5em minus 0.4em\relax The MIT Press,
  2005.

\bibitem{2016arXiv160803140W}
W.~Wan \emph{et~al.}, ``{A Mid-level Planning System for Object
  Reorientation},'' \emph{ArXiv e-prints}, 2016.

\end{thebibliography}
\end{document}